\documentclass[10pt,twocolumn,letterpaper]{article}

\usepackage{cvpr}              % To produce the CAMERA-READY version
\usepackage[dvipsnames]{xcolor}

\definecolor{cvprblue}{rgb}{0.21,0.49,0.74}
\definecolor{asparagus}{rgb}{0.53, 0.66, 0.42}
\definecolor{airforceblue}{rgb}{0.36, 0.54, 0.66}
\usepackage[pagebackref,breaklinks,colorlinks,citecolor=cvprblue]{hyperref}

\usepackage[capitalize]{cleveref}
\crefname{section}{Sec.}{Secs.}
\Crefname{section}{Section}{Sections}
\Crefname{table}{Table}{Tables}
\crefname{table}{Tab.}{Tabs.}

\newcommand\ch[1]{\textcolor{black}{#1}} % chaerin
\newcommand\sh[1]{\textcolor{black}{#1}} % soohyeok
\newcommand\ws[1]{\textcolor{black}{#1}} % wonsuk

\newcommand\blfootnote[1]{%
  \begingroup
  \renewcommand\thefootnote{}\footnote{#1}%
  \addtocounter{footnote}{-1}%
  \endgroup
}

\def\paperID{5} % *** Enter the Paper ID here
\def\confName{CVPR}
\def\confYear{2024}

\title{Fashion Style Editing with Generative Human Prior}

\author{Chaerin Kong$^*$, Seungyong Lee$^*$, Soohyeok Im$^*$, Wonsuk Yang$^*$ \\
NXN Labs \\
\texttt{\{chaerin,seungyong,soohyeok,wonsuk\}@nxn.ai}
}

\begin{document}
\twocolumn[{
\renewcommand\twocolumn[1][]{#1}
\maketitle

% -----------------------------------
% fig 1. main figure
\begin{center}
    \centering
    \captionsetup{type=figure}
    \includegraphics[width=\textwidth]{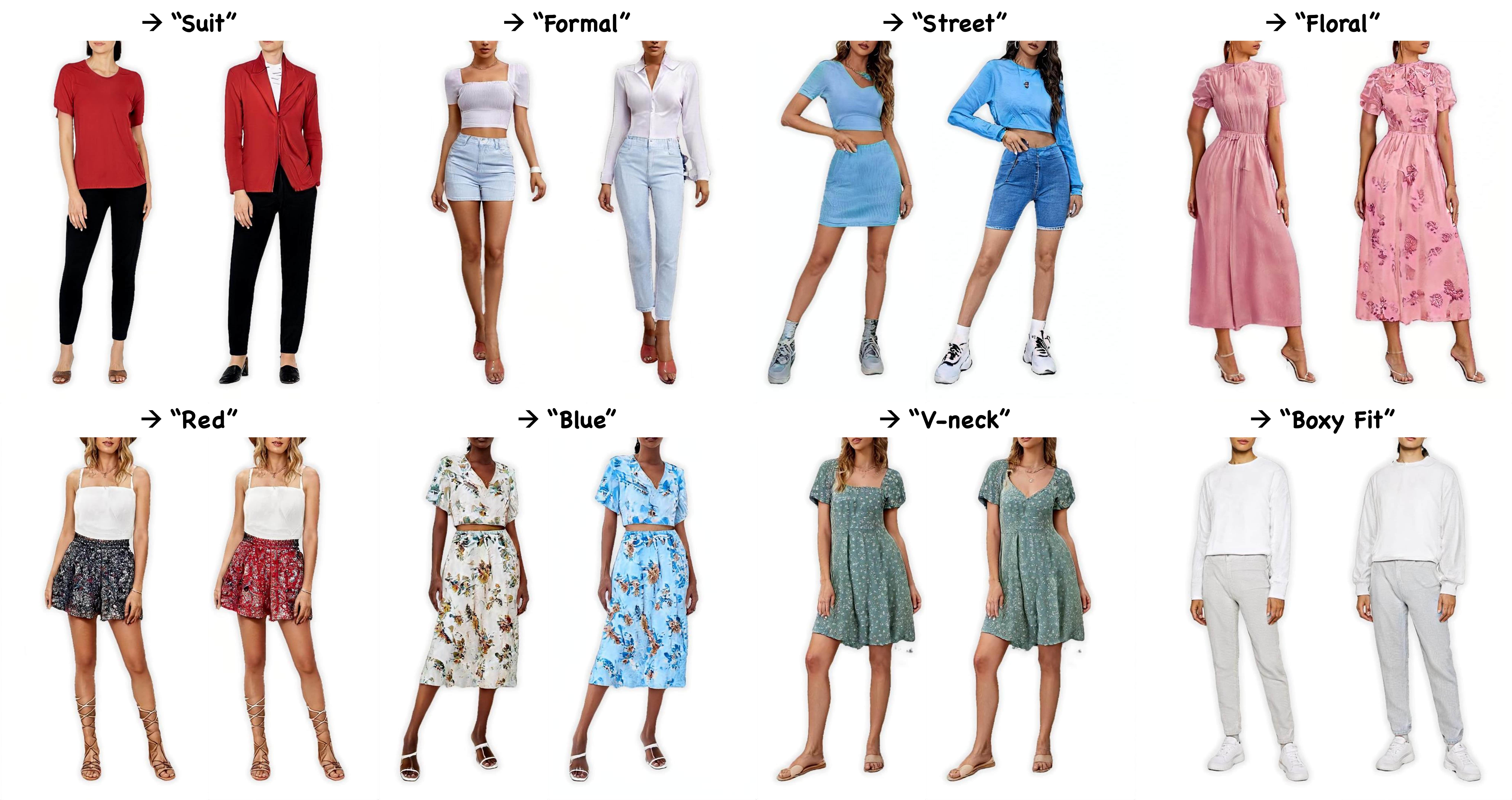}
    \captionof{figure}{Example of text-driven fashion style edition using our framework (FaSE). Given an input image, a text prompt is used to drive change in garment style while preserving its overall attributes.}
    \label{fig:main}
\end{center}
% -----------------------------------

}]
\begin{abstract}
Image editing has been a long-standing challenge in the research community with its far-reaching impact on numerous applications. 
Recently, text-driven methods started to deliver promising results in domains like human faces, but their applications to more complex domains have been relatively limited.
In this work, we explore the task of fashion style editing, where we aim to manipulate the fashion style of human imagery using text descriptions. 
\blfootnote{
$^*$ Equal contributions.
}
% We first investigate how the existing editing framework can be tailored to our application, and propose modifications to enable visualization of abstract concept in the fashion domain. 
Specifically, we leverage a generative human prior and achieve fashion style editing by navigating its learned latent space. 
We first verify that the existing text-driven editing methods fall short for our problem due to their overly simplified guidance signal, and propose two directions to reinforce the guidance: textual augmentation and visual referencing. 
Combined with our empirical findings on the latent space structure, our \textbf{Fa}shion \textbf{S}tyle \textbf{E}diting framework (\textbf{FaSE}) successfully projects abstract fashion concepts onto human images and introduces exciting new applications to the field.
\end{abstract}

\section{Introduction}
\label{sec:int}

\ch{
Recent years have witnessed unprecedented progress in generative models~\cite{ho2020denoising, dhariwal2021diffusion, sauer2023stylegan, kong2022few, ho2022video}, with the language guidance (\textit{i.e.,} text-to-image) being the cornerstone for their enhanced controllability and generalizability~\cite{ramesh2021zero, saharia2022photorealistic, rombach2022high, bar2024lumiere, singer2022make, lee2023aadiff, kong2023analyzing}. Image editing task~\cite{kwon2022clipstyler, shen2021closed, kong2023leveraging} also started to draw great attention from the public thanks to the intuitive editing interface text-to-image frameworks provide~\cite{avrahami2022blended, patashnik2021styleclip, hertz2022prompt}. One of the most popular ways to incorporate text condition to the image editing pipeline is to use a pretrained vision-language foundation model~\cite{radford2021learning, li2022blip, yu2022coca, alayrac2022flamingo, jang2023unifying, jang2023self} as in StyleCLIP~\cite{patashnik2021styleclip}, CLIPstyler~\cite{kwon2022clipstyler} or diffusionCLIP~\cite{kim2022diffusionclip}, where a joint embedding model like CLIP~\cite{radford2021learning} guides the editing direction in the form of similarity score. Although this approach has demonstrated impressive editing capability in some visual domains like human faces~\cite{karras2019style}, they are vulnerable to the blindspots of CLIP features. Specifically, CLIP exposes weakness in domain-specific applications that requires special knowledge~\cite{wang2022medclip, chia2022contrastive}, and its tendency to focus on the global semantic often renders it unsuitable for guiding local editing operation, especially when the image contains complex semantics~\cite{kong2023analyzing}.
}

% \ch{
% CLIP, though powerful and robust, has several limitations that hinders its usage in image editing scenarios. First, it is a general purpose model that is designed to embody the world knowledge in a balanced manner. 
% % While this nature positions it as an ideal foundation for a broad spectrum of applications, leveraging it in highly specialized domains without additional tuning typically exposes weakness~\cite{wang2022medclip, chia2022contrastive}. 
% Second, it abstracts both visual and textual signal into a single \texttt{[CLS]} token, which encourages the learned feature to encode the global discriminative semantic rather than local details and scene composition~\cite{zhong2022regionclip, li2022clip}. 
% % This renders CLIP signal unsuitable for local editing task, especially when the image contains complex semantics~\cite{kong2023analyzing}.
% }

\ch{
Unfortunately, these vulnerabilities are greatly highlighted in our problem setting: \textit{fashion style editing of human imagery}. 
% Although generative modeling in fashion domain has been studied from multiple angles including garment design~\cite{yu2019personalized, sun2023sgdiff}, virtual try-on~\cite{choi2021viton, xu2024ootdiffusion, kim2023stableviton} and outfit recommendation~\cite{lu2021personalized, ding2023personalized}, an 
In an attempt to introduce an intuitive interface for human fashion restyling,
we explore the creative task of text-driven fashion style editing, where we steer the \textit{style} of full-body human portrait with language guidance such as ``street fashion" or ``floral pattern" (\cref{fig:main}). With our unique problem scope arises a new set of challenges, namely the modeling complexity of whole human body imagery and the elusiveness of fashion concepts, \textit{i.e.,} how some terms like \textit{`street fashion'} can induce different mental images to individuals.
% The former has been discussed by the authors of StyleGAN-Human~\cite{fu2022styleganhuman}, a pioneering work in human image generation, while the latter is clear if we think about how individuals could have different mental images towards high-level fashion concepts like \textit{preppy} or \textit{street}.
These two distinguish our approach from previous endeavors that tried to manipulate human faces with text guidance~\cite{patashnik2021styleclip, kwon2022clipstyler}, in that we need to explicitly reformulate our guidance (\textit{i.e.,} fashion style) into a more illustrative and visually grounded learning signal.
}

\ch{
To this end, we propose two directions to visually reinforce the guidance signal. First, we enrich the text prompt by querying a language model~\cite{touvron2023llama, brown2020language} to generate descriptions for the concept. Second, we retrieve reference images for the target text from a fashion database we constructed through web crawling, and guide our model to consult them for more descriptive guidance. Leveraging the enhanced visual clarity, our \textbf{Fa}shion \textbf{S}tyle \textbf{E}diting (\textbf{FaSE}) framework adeptly transforms abstract fashion concepts into tangible human imagery across a variety of applications.
% todo: make guidance visually clear and illustrative
% A new set of challenges arises in our problem scope, namely the modeling complexity of whole human body imagery and the elusiveness of fashion concepts. 
% significance of our task, 
% nature of our task - whole human (complex), fashion specific concepts (less descriptive)
% Direct application of CLIP guidance (ex StyleCLIP) yields suboptimal results
}

% \ch{
% To this end, we propose to clarify/reinforce the guidance signal
% Textual augmentation
% Visual referencing
% For application, tackle hierarchical latent space
% }

\section{Related Work}
\label{sec:rel}
\noindent
\ch{
\textbf{Image editing} poses unique technical challenges stemming from the famous distortion-editability trade-off, \textit{i.e.,} making necessary changes while preserving the original. The most prominent approach is leveraging a pretrained generative model to constrain the output to the learned data manifold and executing edits in its latent space. StyleGAN family~\cite{karras2019style, karras2020analyzing, karras2021alias, karras2020training} has been a popular choice for the generative prior for long, thanks to its remarkable generation quality and well-behaved latent space~\cite{wu2021stylespace, lang2021explaining}. Many works~\cite{patashnik2021styleclip, tov2021designing, richardson2021encoding, shen2021closed, kwon2022clipstyler, chai2021using} have demonstrated impressive editing performance by exploring its learned latent space, but their application was mainly confined to narrow semantic domains (\textit{e.g.,} human faces~\cite{karras2019style}) that can be readily steered by simple words (\textit{e.g., ``smile"})~\cite{patashnik2021styleclip, richardson2021encoding}. Recently, diffusion-based image editing~\cite{kim2022diffusionclip, meng2021sdedit, kwon2022diffusion, zhang2023sine, hertz2022prompt, kong2023leveraging} is also being widely explored.
\newline
\textbf{Human generation} fundamentally differs from human face modeling as the underlying complexity is much greater to synthesize realistic rendering of whole human body with natural garments, body proportions and poses. StyleGAN-Human~\cite{fu2022styleganhuman} pioneered this direction of research from a data-centric perspective, with following works~\cite{zhang2022humandiffusion, fu2023unitedhuman, jiang2022text2human} further pushing the boundary. In our work, this learned prior serves as the starting point for fashion-specific style editing. 
\newline
\textbf{Fashion image generation}, due to its high practical impact, has been studied from various angles, such as virtual try-on~\cite{choi2021viton, xu2024ootdiffusion, morelli2022dress, lee2022high, li2023virtual}, clothing synthesis~\cite{jiang2017fashion, rostamzadeh2018fashion} and clothing editing~\cite{kong2023leveraging}. However, manipulating human-fitted fashion imagery is under-explored due to its innate complexity. FE-GAN~\cite{dong2020fashion} proposes an image-based editing framework that relies on adversarial training, but the editing operation is confined to few cases (\textit{e.g.,} sleeve length) and the results are generally unsatisfactory. VPTNet~\cite{kwon2022tailor} introduces an attribute-based editing technique, which suffers from similar drawback that only limited operations (sleeve and length) predefined at the training phase are supported. In this work, we aim to steer the fashion \textit{style} of human imagery with high-level user input keywords (\textit{e.g.,} ``formal") in a highly versatile and compute-efficient manner.
}

% -----------------------------------
% fig 2. clip loss
\begin{figure}[t]
\centering
\begin{subfigure}{.23\textwidth}
  \centering
  \includegraphics[width=1.\linewidth]{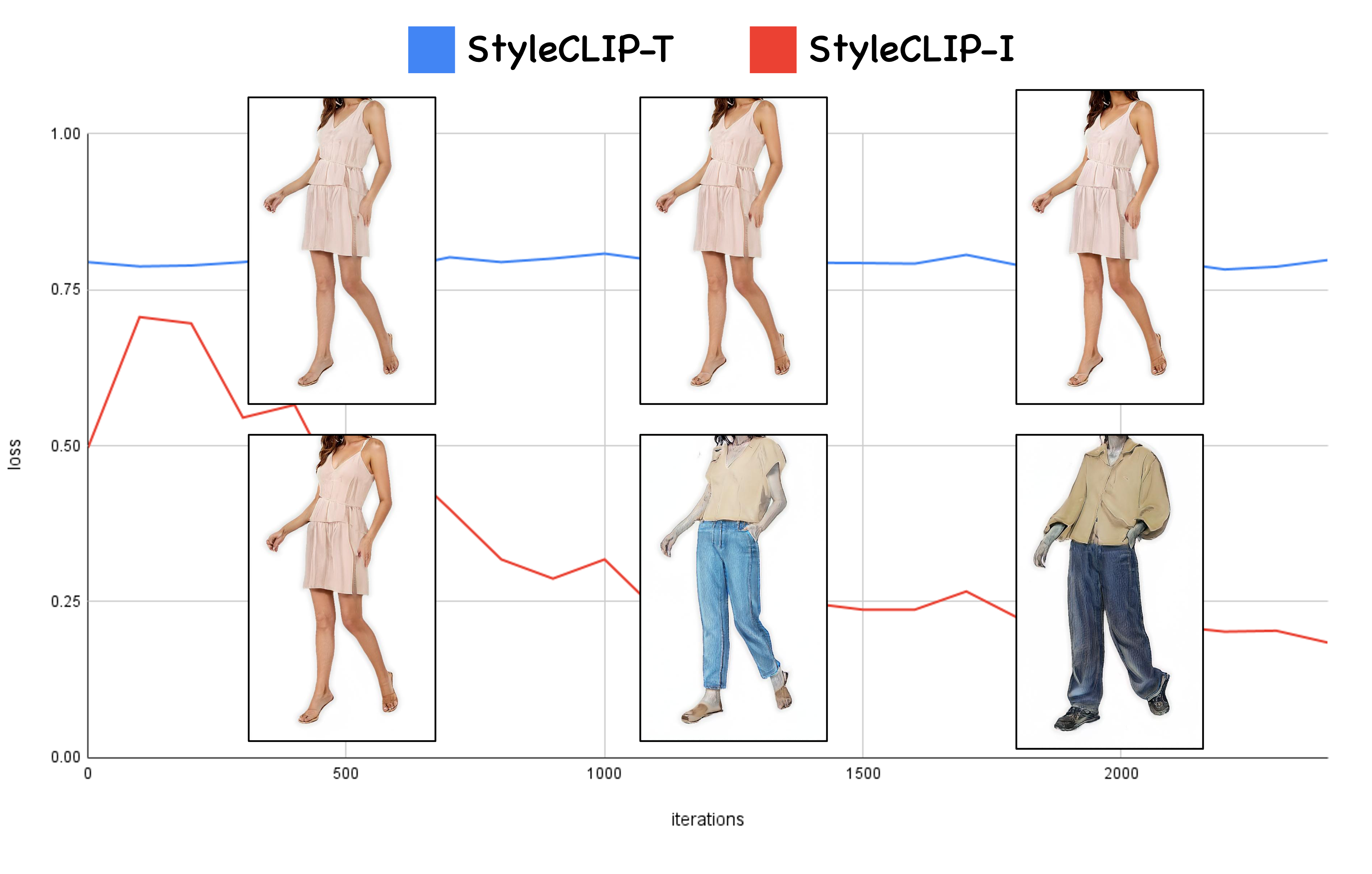}
  \caption{Naive CLIP guidance.} 
  \label{fig:loss1}
\end{subfigure}%
\begin{subfigure}{.23\textwidth}
  \centering
  \includegraphics[width=1.\linewidth]{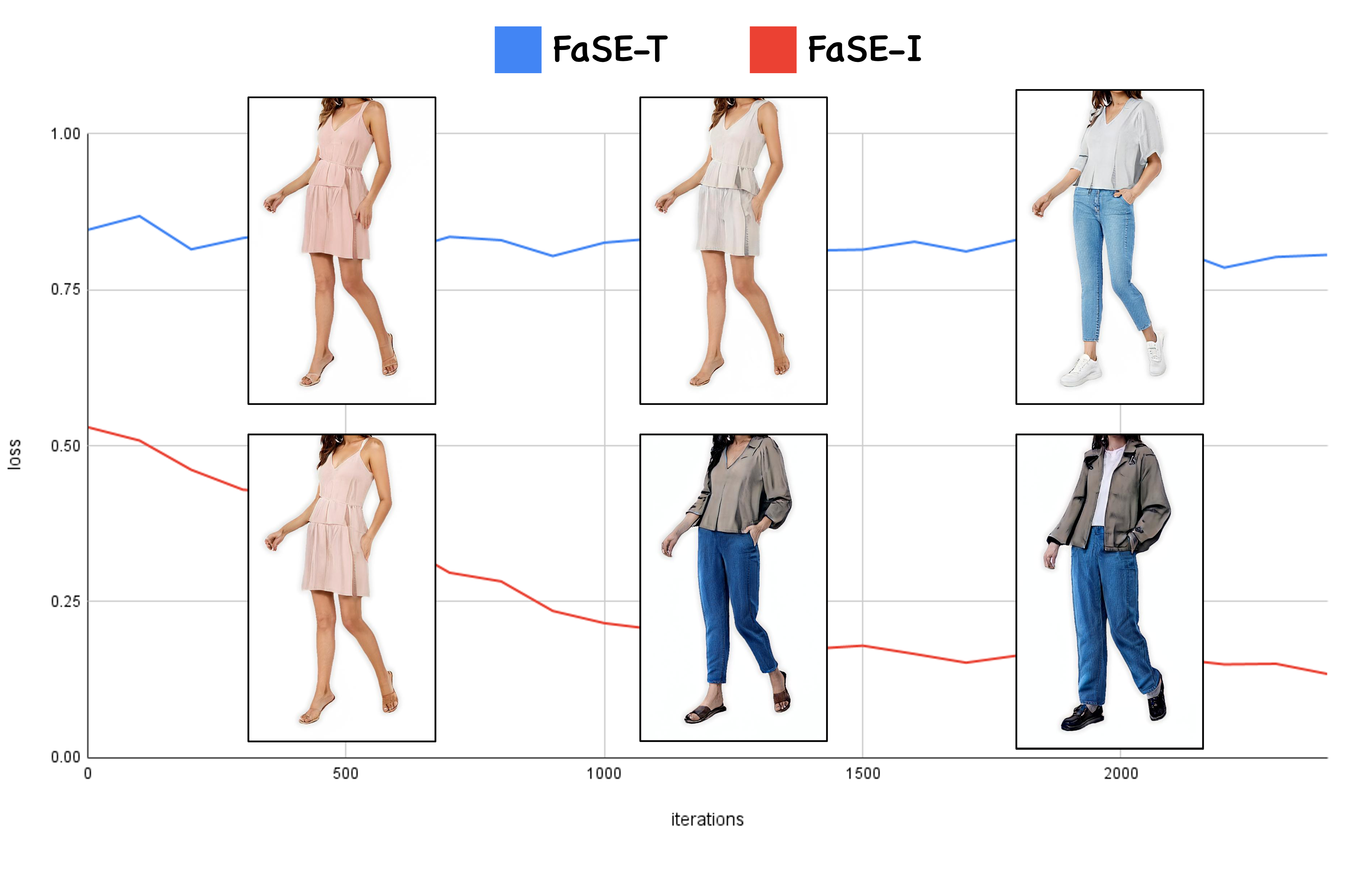}
  \caption{Descriptive guidance (FaSE).}
  \label{fig:loss2}
\end{subfigure}
\caption{Using CLIP text guidance as in StyleCLIP~\cite{patashnik2021styleclip} is insufficient to steer fashion concepts in full-body images, and image guidance in CLIP feature space often distorts the semantics (\textit{left}). FaSE, in contrast, successfully projects non-trivial fashion styles (\textit{`street-fashion'} in this example) onto human images with both type of guidances (\textit{right}).}
\label{fig:loss}
\end{figure}
% -----------------------------------
% -----------------------------------
% figure 3 -- model architecture
\begin{figure*}
\centering
\includegraphics[width=.8\textwidth]{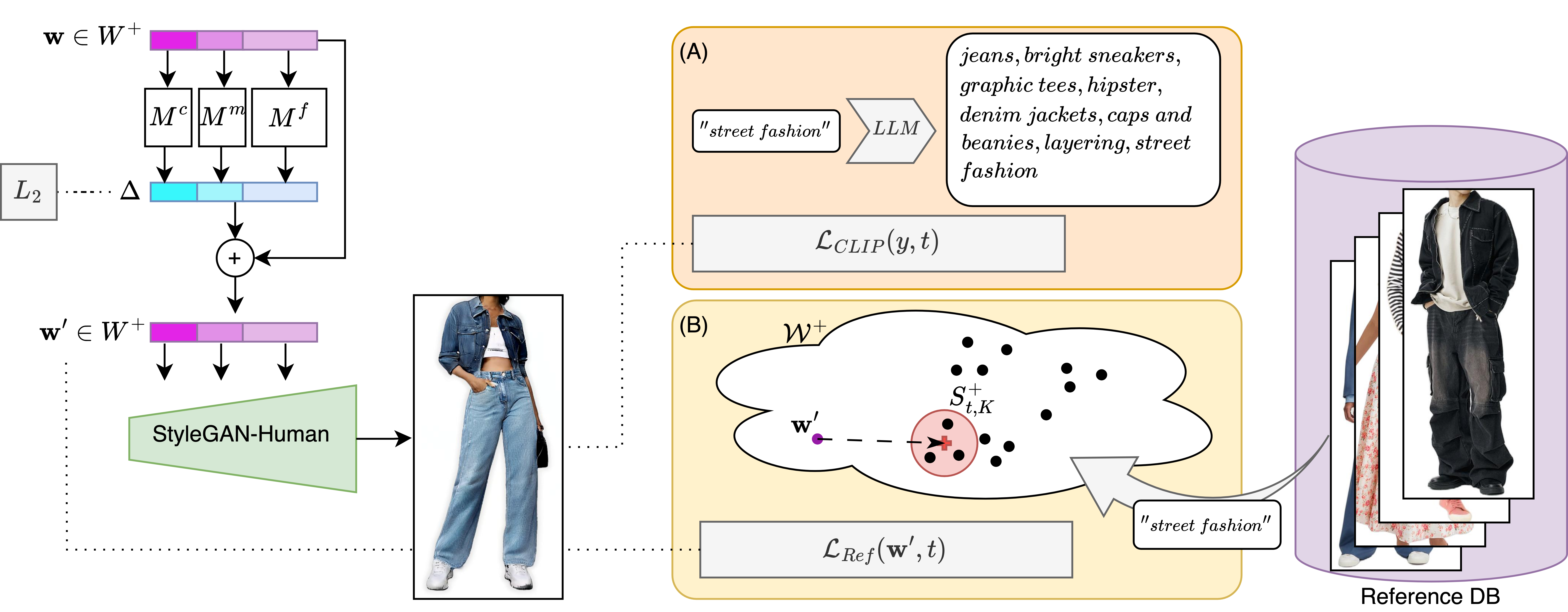}
\caption{Summary of our fashion style editing framework (FaSE). We learn a latent mapper~\cite{patashnik2021styleclip} with two types of illustrative guidance. (A) We transform fashion style concepts into visual descriptions with a pretrained language model. (B) From our preconstructed fashion image database, we retrieve top-k reference images that both suit the target text prompt and resemble the source image, which are used as additional visual guidance.
% We encourage the source image to absorb key visual concepts of the reference image.
}
\label{fig:model}
\end{figure*}
% -----------------------------------

\section{Approach}
\label{sec:app}

\subsection{Preliminaries}
\label{subsec:pre}
\noindent
\ws{
Our framework is based on StyleGAN-Human \cite{fu2022styleganhuman}, which serves as a pretrained prior for human image generation. For editing, we employ the latent mappers introduced in StyleCLIP \cite{patashnik2021styleclip}. Latent mappers are fully-connected networks that transform any latent vector in StyleGAN's $\mathcal{W}+$ space to the direction that better aligns with the provided text prompt. Specifically, the latent vector is split into three parts (coarse, medium, and fine), each responsible for different level of detail. Each part undergoes mapping through distinct networks ($M^c$, $M^m$, $M^f$, respectively) and is subsequently added to the original latent vector before being fed to the generator. The network is trained to minimize the cosine distance between the generated image and the text prompt in CLIP joint embedding space along with an $L_2$ regularization to preserve the original image.
}
\ch{
Although it has delivered impressive results in human face editing, its coarse guidance signal is insufficient for controlling non-trivial fashion styles on full-body human images (\cref{fig:loss}).
}

\subsection{Textual Augmentation}
\label{subsec:tex}
\noindent
\sh{
The key to successful fashion style editing lies in providing \ch{specific visual} guidance that represents the intended concept. 
\ch{As StyleCLIP utilizes generic CLIP loss with a naive text prompt $t$, training often fails for non-trivial fashion concepts, as shown in \cref{fig:loss}.}
% While StyleCLIP utilizes CLIP loss with a simple \emph{text prompt} $t$, it fails to edit the image as desired if the given concept is not straightforward. 
% We found that the failure 
% \ch{can be attributed}
% of the naïve approach is due 
% to the lack of expressiveness in the text prompt.
\ch{We hypothesize that this failure can be attributed to the entanglement in StyleGAN-Human latent space and the innate elusiveness of fashion terminologies.
A straightforward yet effective solution would be enriching the text prompt with detailed descriptions, thereby enabling its decomposition into specific components that correspond to various visual aspects of the concept.
}
% In the fashion domain, there are often complementary or sub-items, making it impossible for a single word to fully encompass a specific concept. 
\ch{For example, in visualizing the concept of `formal fashion', identifying specific components like `suit,' `trouser,' and `loafer' proves to be beneficial. 
To this end, 
}
% To address this comprehensive property, 
we leverage a Large Language Model (LLM) to augment the text prompt $t$, making the context more illustrative by concatenating component words associated with the given concept. 
\ch{Formally, we train the latent mapper to minimize CLIP loss with respect to the augmented text prompt as follows:}
% Then, low to high-level fashion concepts can be trained by minimizing CLIP loss with an expanded text prompt
\[
    \mathcal{L}_{CLIP}(y, t) = \mathcal{D}(\mathcal{E}_{CLIP}^{img}(y), \mathcal{E}_{CLIP}^{text}(g(t)))
\]
where $\mathcal{D}$ denotes the cosine distance, $\mathcal{E}_{CLIP}$ denotes the CLIP encoder, \ch{$g$ stands for the textual augmentation operation using LLM} and $y$ is the edited image. 
}

\subsection{Editing with Visual Reference}
\label{subsec:ret}
\noindent
\ch{
Upon observing the effectiveness of textual augmentation, a natural progression would be utilizing visual references corresponding to the target concept, since \textit{an image is worth a thousand words}. 
To this end, we first construct a reference database by collecting clean human images corresponding to various fashion style concepts from the web. 
Our high level idea is to retrieve an array of relevant images and train our model not only to increase the cross-modal CLIP similarity but also to maximize the image-level similarity with the reference set.
Our initial investigation into the CLIP image space reveals its unsuitability for our task, prompting us to transition into a more illustrative feature space, \textit{i.e.,} the $\mathcal{W}+$ space~\cite{patashnik2021styleclip} of the generator. Formally, our objective can be written as:
\[
    \mathcal{L}_{Ref}(\mathbf{w'}, t) = \frac{1}{K} \sum_{s \in S_{t, K}^+} \mathcal{D}(\mathbf{w'}, s),
\]
where $\mathbf{w}^\prime$ is the translated latent vector and $S_{t, K}^+$ is the collection of $\mathcal{W}+$ embeddings of top-K reference images based on similarity.
}

\ch{
As mentioned earlier, we empirically find that naively optimizing image similarity in CLIP image space leads to sub-optimal results with undesired distortions, mainly because CLIP visual features are too coarse to capture the subtle nuances of fashion. Hence, we first invert the retrieved reference images to $\mathcal{W}+$ space and train the mapper to push latent codes towards them, injecting a more direct and illustrative learning signal. Top-K retrieval is performed to suppress irrelevant guidance coming from overly dissimilar reference images.
}

\subsection{Navigating the Hierarchical Latent Space}
\label{subsec:hie}
% \sy{
% In StyleGAN2 \cite{karras2021alias}, the authors demonstrated a hierarchy in style embedding from global to local features. Inspired by this, in StyleCLIP \cite{patashnik2021styleclip}, the authors edited human faces by modifying 18 style embeddings using a latent mapper, divided hierarchically into coarse (from 1st to 4th embeddings), medium (from 5th to 8th embeddings), and fine (from 9th to the last embeddings). We leverage this hierarchical structure in the latent space ($\mathcal{W}^+$) of StyleGAN-Human, known as a generative human prior model, to enable selective human fashion editing.
% \newline
% As illustrated in Fig. \ref{fig:coarse}, using $M^c$ or $M^f$ to modify the coarse or fine embeddings, respectively, makes it difficult to alter the type of garments. However, using $M^m$ to change the medium embeddings makes it possible to modify the type of garments. Similarly, colors and patterns of garments can be effectively edited by modifying fine embeddings through $M^f$. However, modifying coarse embeddings through $M^c$ is not very effective for fashion editing because it controls non-fashion features such as human pose.
% }
\ch{
StyleGAN2~\cite{karras2020analyzing} is known for its hierarchical latent space, which has motivated many applications that leverage its coarse-to-fine structure. StyleCLIP~\cite{patashnik2021styleclip}, for example, divides the latent space into three groups and empirically discovered which level corresponds to specific facial editing operation. We look for a similar empirical guideline for fashion editing on StyleGAN-generated human images. 
}

\begin{figure}
    \centering
    \includegraphics[width=0.9\linewidth]{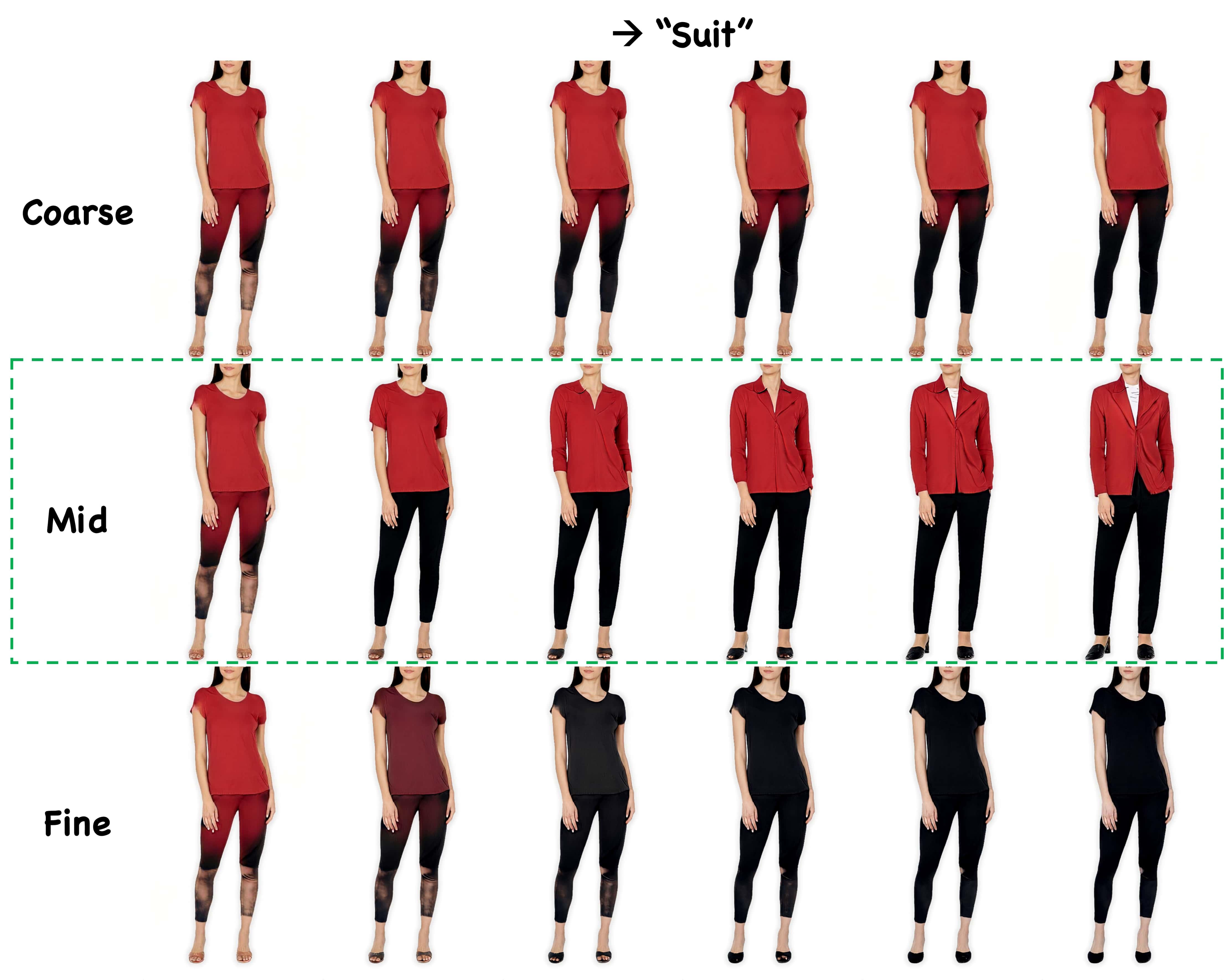}
\textbf{}    \caption{From the top, we train $M^c$, $M^m$, and $M^f$, respectively, for the prompt \textit{`suit'}. When editing the garment shape as in \textit{`suit'}, $M^m$ needs to be modified to transform the mid part of $\mathbf{w}$.}
    \label{fig:coarse}
\end{figure}

\ch{
Following \cite{patashnik2021styleclip}, we first compartmentalize the $\mathcal{W}+$ space into three levels, and test editing operations of various granularity, \textit{e.g.,} changing the color, neckline, garment category, etc. We discover that the coarse group has little association with fashion as it is mainly responsible for human pose, and mid/fine group controls shape/texture respectively. From \cref{fig:coarse}, for example, it is evident that in order to modify garment category, mid-level transformation is necessary.
}

\section{Experiments}
\label{sec:exp}

\ch{
We choose StyleCLIP with StyleGAN-Human-v2 checkpoint\footnote{https://github.com/stylegan-human/StyleGAN-Human} as our baseline. 
To construct the reference DB, we collect fashion images from online marketplaces. We select 12 categories (\textit{e.g.,} boxy fit, floral pattern, street fashion, etc) and gather 35 images for each category. 
}

\subsection{Qualitative Results}
\label{subsec:qual}
\begin{figure}
    \centering
    \includegraphics[width=0.9\linewidth]{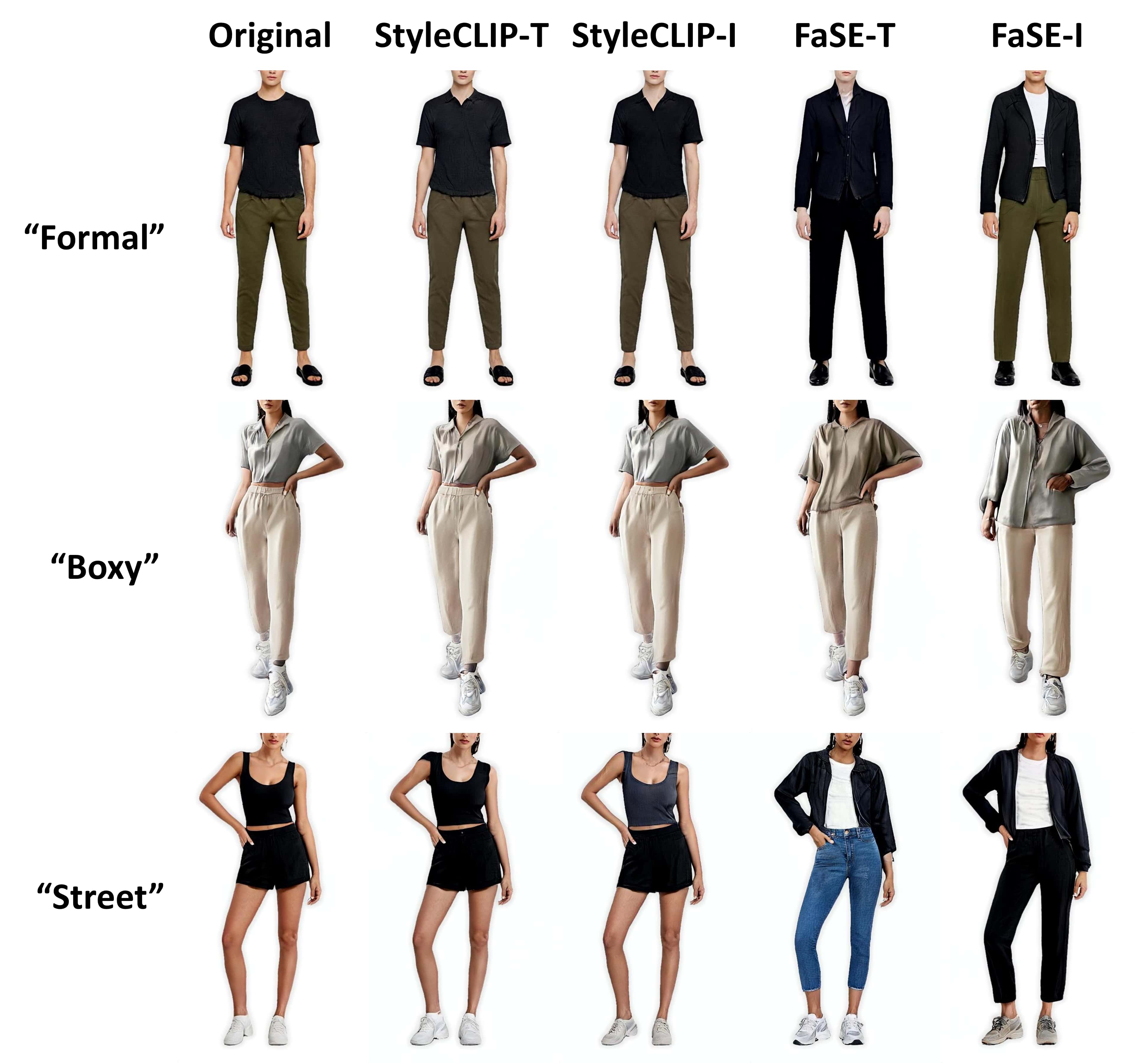}
    \caption{Qualitative comparison with StyleCLIP using text and image guidance signal.}
    \label{fig:qual}
\end{figure}
% -----------------------------------

\ch{
We first qualitatively compare our method against StyleCLIP baseline using both text (StyleCLIP-T, FaSE-T) and image (StyleCLIP-I, FaSE-I) signal. We can see from \cref{fig:qual} that while naive CLIP signal, whether it be textual or visual, fails to modify human images, ours successfully alters them towards the target direction. FaSE-I, in particular, incurs little change to the original image semantics thanks to the retrieval mechanism.
}

% table 1: GPT4v 
\begin{table}[t]
    \centering
    \footnotesize
    \begin{tabular}{lccccc}
    \specialrule{.1em}{.05em}{.05em}
    & \multicolumn{2}{c}{GPT-4V} & \multicolumn{2}{c}{User Study} \\
    Prompt & \textbf{FaSE-T} & \textbf{FaSE-I} & \textbf{FaSE-T} & \textbf{FaSE-I}\\
    \hline
    \textbf{T-shirts} & 55.9 & 58.8 & 56.9 & 52.0 \\
    \textbf{Floral} & 72.5 & 67.6 & 63.7 & 62.7 \\
    \textbf{Boxy} & 91.2 & 73.5 & 61.8 & 76.5 \\
    \textbf{Formal} & 91.2 & 96.1 & 90.2 & 94.1 \\
    \textbf{Street} & 95.1 & 88.2 & 79.4 & 80.4 \\
    \specialrule{.1em}{.05em}{.05em}
    \end{tabular}
    \caption{Text-image alignment using GPT-4V and user test. We report 1v1 win rate against \cite{patashnik2021styleclip} on diverse text prompts.}
    \label{tab:quan}
\end{table}
% -----------------------------------
\subsection{Quantitative Results}
\label{subsec:quan}

For the quantitative evaluation, we report the 1v1 win rate as assessed by GPT4-V and human raters. For GPT4-V evaluation, we prompt the model with the template: ``\textit{Between the two images, which do you think better aligns with the fashion concept `[concept name]'?}" \cref{tab:quan} clearly shows that our methods have comparative advantage over the baseline across various text prompts.

% -----------------------------------
% table 2: user study 
% \begin{table}[t]
%     \centering
%     \footnotesize
%     \begin{tabular}{lcc}
%     \specialrule{.1em}{.05em}{.05em}
%     Method & Result Quality & Prompt Fidelity \\
%     \hline
%     Baseline \cite{patashnik2021styleclip} &  a & b\\
%     \textbf{FaSE} (text-aug.) & c & d \\
%     \textbf{FaSE} (visual-ref.) & e & f \\
%     \specialrule{.1em}{.05em}{.05em}
%     \end{tabular}
%     \caption{Comparison of }
%     \label{tab:quan}
% \end{table}
% -----------------------------------
\section{Conclusion}
\label{sec:con}

\ch {
Our work extended text-driven image editing to fashion style editing by exploring two techniques that visually clarify the guidance signal. We hope our framework facilitates new and exciting applications in fashion field.
}

{
    \small
    \bibliographystyle{ieeenat_fullname}
    \bibliography{main}
}


\end{document}